\documentclass[10pt, conference]{IEEEtran}

\usepackage{epsfig}
\usepackage{graphicx}
\usepackage{amsmath}
\usepackage{amssymb}
\usepackage{microtype}
\usepackage{xcolor}

\usepackage{graphicx}
\usepackage[font=small]{caption}
\usepackage{subcaption}
\graphicspath{{figs/}}

\usepackage{stackengine}
\newcommand{\inlinelabel}[2]{
    \stackinset{l}{-2pt}{b}{-2pt}{\colorbox{white}{\small(#1)}}{#2}
}

\usepackage{paralist}

\usepackage[numbers,square]{natbib}
\usepackage[pagebackref=false,breaklinks=true,bookmarks=false]{hyperref}

\hypersetup{
  pdftitle={anonymous},
  pdfauthor={anonymous},
  bookmarksnumbered,
  pdfstartview={FitH},
  colorlinks,
  citecolor=black,
  filecolor=black,
  linkcolor=black,
  urlcolor=black,
  breaklinks=true,
}

\usepackage{xspace}
\newcommand{\dataset}{SynthesEyes\xspace}

\usepackage{authblk}

\begin{document}

\title{Rendering of Eyes for Eye-Shape Registration \\ and Gaze Estimation}

\author[1]{Erroll Wood\thanks{A.A@university.edu}}
\author[1]{Tadas Baltru\v{s}aitis\thanks{C.C@university.edu}}
\author[2]{Xucong Zhang\thanks{B.B@university.edu}}
\author[2]{Yusuke Sugano\thanks{D.D@university.edu}}
\author[1]{Peter Robinson\thanks{E.E@university.edu}}
\author[2]{Andreas Bulling\thanks{E.E@university.edu}}
\affil[1]{University of Cambridge, United Kingdom
  \texttt{\small\{eww23,tb346,pr10\}@cam.ac.uk}}
\affil[2]{Max Planck Institute for Informatics, Germany
  \texttt{\small\{xczhang,sugano,bulling\}@mpi-inf.mpg.de}}

\maketitle

\begin{abstract}
Images of the eye are key in several computer vision problems, such as shape registration and gaze estimation.
Recent large-scale supervised methods for these problems require time-consuming data collection and manual annotation, which can be unreliable.
We propose synthesizing perfectly labelled photo-realistic training data in a fraction of the time.
We used computer graphics techniques to build a collection of dynamic eye-region models from head scan geometry.
These were randomly posed to synthesize close-up eye images for a wide range of head poses, gaze directions, and illumination conditions.
We used our model's controllability to verify the importance of realistic illumination and shape variations in eye-region training data.
Finally, we demonstrate the benefits of our synthesized training data (\dataset) by out-performing state-of-the-art methods for eye-shape registration as well as cross-dataset appearance-based gaze estimation in the wild.

\end{abstract}

\section{Introduction}

The eyes and their movements convey our attention and play a role in communicating social and emotional information \cite{Argyle1965}.
Therefore they are important for a range of applications including gaze-based human-computer interaction~\cite{majaranta14_apc}, visual behavior monitoring~\cite{bulling11_pami}, and -- more recently -- collaborative human-computer vision systems~\cite{papadopoulos2014training,sattar15_cvpr}. 
Typical computer vision tasks involving the eye include \emph{gaze estimation}: determining where someone is looking, and \emph{eye-shape registration}: detecting anatomical landmarks of the eye, often as part of the face (e.g. eyelids).

Machine learning methods that leverage large amounts of training data currently perform best for many problems in computer vision, such as object detection~\cite{girshick2014rich}, scene recognition~\cite{zhou2014learning}, or gaze estimation~\cite{zhang15_cvpr}.
However, capturing data for supervised learning can be time-consuming and require accurate ground truth annotation.
This annotation process can be expensive and tedious, and there is no guarantee that human-provided labels will be correct.
Ground truth annotation is particularly challenging and error-prone for learning tasks that require accurate labels, such as tracking facial landmarks for expression analysis, and gaze estimation.

\begin{figure}
    \includegraphics[width=\columnwidth]{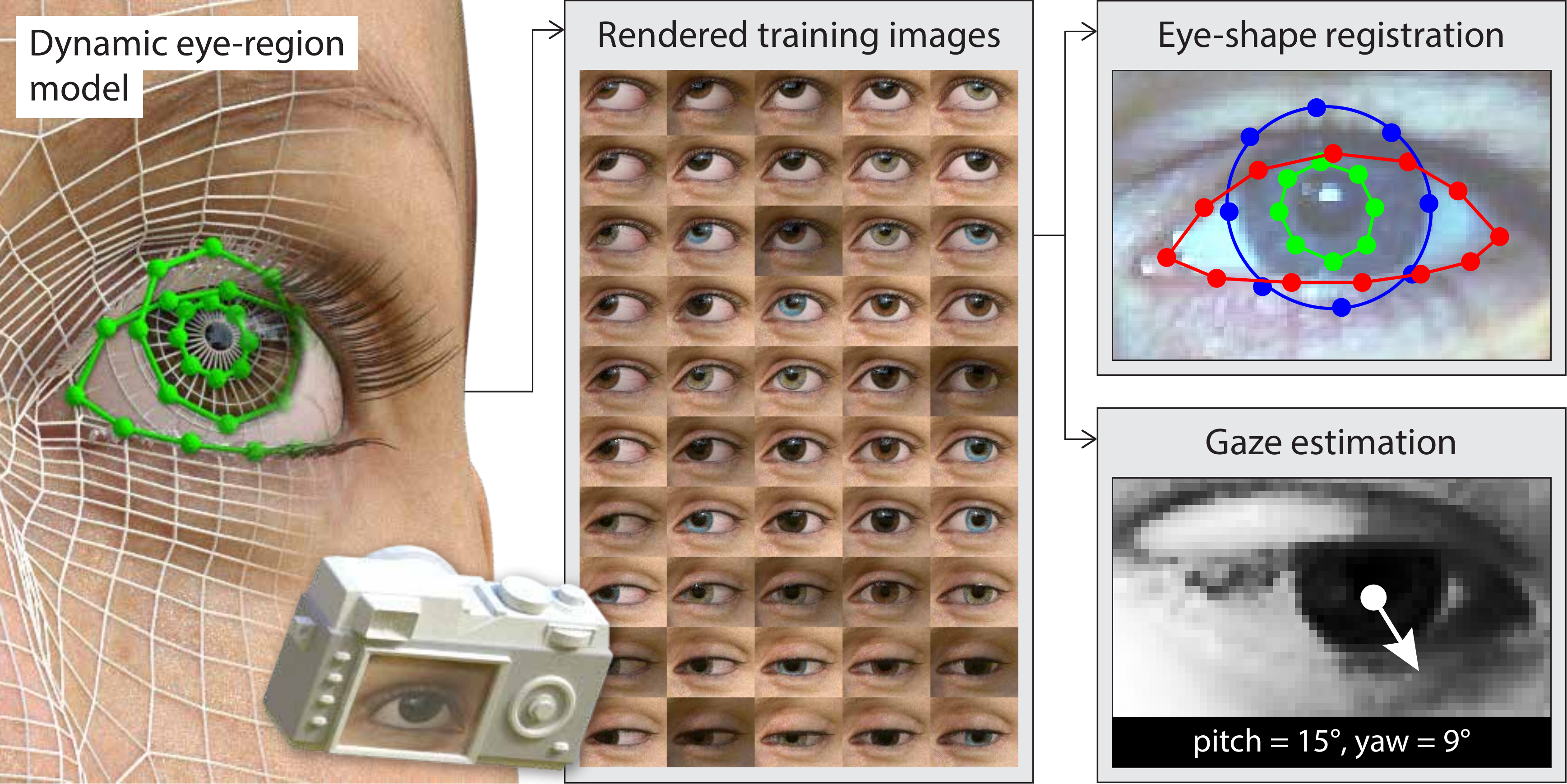}
    \caption{We render a large number of photorealistic images of eyes using a dynamic eye region model. These are used as training data for eye-shape registration and appearance-based gaze estimation.}
    \label{fig:teaser}
\end{figure}

To address these problems, researchers have employed \emph{learning-by-synthesis} techniques to generate large amounts training data with computer graphics.
The advantages of this approach are that both data collection and annotation require little human labour and image synthesis can be geared to specific application scenarios.
The eye-region is particularly difficult to model accurately given the dynamic shape changes it undergoes with facial motion and eyeball rotation, and the complex material structure of the eyeball itself.
For this reason, recent work on learning-by-synthesis for gaze estimation employed only fundamental computer graphics techniques -- rendering low-resolution meshes without modeling illumination changes or accounting for the varying material properties of the face \cite{sugano2014learning}.
In addition, these models are not fully controllable and the synthesized datasets contain only gaze labels, limiting their usefulness for other computer vision problems, such as facial landmark registration.

%!TEX root = ../00_main.tex

%
% I wanted to keep links while using inset labels to save space
% So empty captions are labelled, and we use a big negative vspace
% This is obviously a hack, but I think it's worthwhile
% - Erroll
%
\begin{figure*}[ht]
    \captionsetup[subfigure]{labelformat=empty} % stop subcaption writing "(a)""
    \captionsetup{subrefformat=parens} % add parentheses to \subref
    \centering
    \begin{subfigure}[t]{0.195\textwidth}
        \inlinelabel{a}{\includegraphics[width=\textwidth]{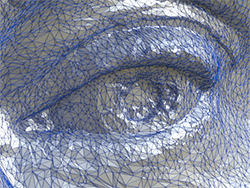}}
        \caption{}\label{fig:process_original_scan}
    \end{subfigure}
    \hfill
    \begin{subfigure}[t]{0.195\textwidth}
        \inlinelabel{b}{\includegraphics[width=\textwidth]{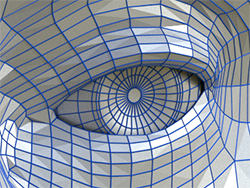}}
        \caption{}\label{fig:process_retopo}
    \end{subfigure}
    \hfill
    \begin{subfigure}[t]{0.195\textwidth}
        \inlinelabel{c}{\includegraphics[width=\textwidth]{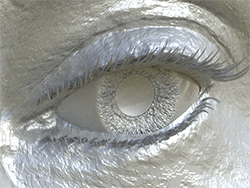}}
        \caption{}\label{fig:process_displaced_subdiv}
    \end{subfigure}
    \hfill
    \begin{subfigure}[t]{0.195\textwidth}
        \inlinelabel{d}{\includegraphics[width=\textwidth]{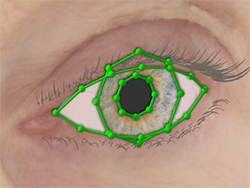}}
        \caption{}\label{fig:process_ldmks}
    \end{subfigure}
    \hfill
    \begin{subfigure}[t]{0.195\textwidth}
        \inlinelabel{e}{\includegraphics[width=\textwidth]{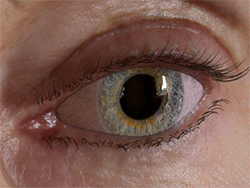}}
        \caption{}\label{fig:process_final_render}
    \end{subfigure}
    \par\vspace{-28pt}
    \caption{An overview of our model preparation process: Dense 3D head scans (1.4 million polygons) \subref{fig:process_original_scan} are first retopologised into an optimal form for animation (9,005 polygons) \subref{fig:process_retopo}. High resolution skin surface details are restored by displacement maps \subref{fig:process_displaced_subdiv}, and 3D iris and eyelid landmarks are annotated manually \subref{fig:process_ldmks}. A sample rendering is shown \subref{fig:process_final_render}.}
    \label{fig:process}
\end{figure*}

We present a novel method for rendering realistic eye-region images at a large scale using a collection of dynamic and controllable eye-region models.
In contrast to previous work, we provide a comprehensive and detailed description of the model preparation process and rendering pipeline (see~\autoref{fig:process} for an overview of the model preparation process and~\autoref{fig:eye_model} for the eye model used).
We then present and evaluate two separate systems trained on the resulting data (\emph{\dataset}): an eye-region specific deformable model and an appearance-based gaze estimator.
The controllability of our model allows us to quickly generate high-quality training data for these two disparate tasks.
Please note that our model is not only limited to these scenarios but can potentially be used for other tasks that require realistic images of eyes, e.g. gaze correction or evaluation of iris-biometrics or geometry-based gaze estimation \cite{swirski2014rendering}.

The specific contributions of this work are threefold.
We first describe in detail our novel but straight-forward techniques for generating large amounts of synthesized training data, including wide degrees of realistic appearance variation using image-based-lighting.
We then demonstrate the usefulness of \dataset by out-performing state-of-the-art methods for eye-shape registration as well as challenging cross-dataset appearance-based gaze estimation in the wild.
Finally, to ensure reproducibility and stimulate research in this area, we will make the eyeball model and generated training data publicly available at time of publication.

\section{Related Work}

Our work is related to previous work on 1) learning using synthetic data and 2) computational modeling of the eyes.

\subsection{Learning Using Synthetic Data}

Despite their success,
the performance of learning-based approaches critically depends on how well the test data distribution is covered by the training set.
Since recording training data that covers the full distribution is challenging,
synthesized training data has been used instead.
Previous work demonstrates such data to be beneficial for tasks such as body pose estimation~\cite{okada2008relevant,shotton2013real}, object detection/recognition~\cite{fu2011neural,yu2010improving,liebelt2010multiview,peng2014exploring}, and facial landmark localization \cite{baltrusaitis20123d,zface}.
Since faces exhibit large color and texture variability, some approaches side-stepped this by relying on depth images \cite{baltrusaitis20123d,fanelli2011real}, and synthesizing depth images of the head using existing datasets or a deformable head-shape model.
Recent work has also synthesized combined color and geometry data by sampling labelled 3D-videos for training a dense 3D facial landmark detector \cite{zface}.

As discussed by \citet{kaneva2011evaluation}, one of the most important factors is the realism of synthesized training images.
If the object of interest is highly complex, like the human eye, it is not clear whether we can rely on overly-simplistic object models.
\citet{zhang15_cvpr} showed that gaze estimation accuracy significantly drops if the test data is from a different environment.
Similarly to facial expression recognition~\cite{stratou2011effect}, illumination effects are a critical factor.
In contrast, our model allows synthesizing realistic lighting effects -- an important degree of variation for performance improvements in eye-shape registration and gaze estimation.

Most similar to this work, \citet{sugano2014learning} used 3D reconstructions of eye regions to synthesize multi-view training data for appearance-based gaze estimation.
One limitation of their work is that they do not provide a parametric model.
Their data is a set of rigid and low-resolution 3D models of eye regions with ground-truth gaze directions, and hence cannot be easily applied to different tasks.
Since our model instead is realistic and fully controllable, it can be used to synthesize close-up eye images with ground-truth eye landmark positions.
This enables us to address eye shape registration via learning-by-synthesis for the first time.

\subsection{Computational Modeling of the Eyes}

The eyeballs are complex organs comprised of multiple layers of tissue, each with different reflectance properties and levels of transparency.
Fortunately, given that realistic eyes are important for many fields, there is already a large body of previous work on modeling and rendering eyes (see Ruhland et al. \cite{ruhland2014look} for a recent survey).

Eyes are important for the entertainment industry, who want to model them with potentially dramatic appearance. \citet{berard2014highquality} represents the state-of-the-art in capturing eye models for actor digital-doubles.
They used a hybrid reconstruction method to separately capture both the transparent corneal surface and diffuse sclera in high detail, and recorded deformations of the eyeball's interior structures. Visually-appealing eyes are also important for the video-game industry. \mbox{\citet{ActiBlizEyes}} recently developed techniques for modeling eye wetness, refraction, and ambient occlusion in a standard rasterization pipeline, showing that approximations are sufficient in many cases.

Aside from visual effects, previous work has used 3D models to examine the eye from a medical perspective.
\citet{sagar1994virtual} built a virtual environment of the eye and surrounding face for mechanically simulating surgery with finite element analysis.
\citet{priamikov14_openeyesim} built a 3D biomechanical model of the eye and its interior muscles to understand the underlying problems of visual perception and motor control.
Eye models have also been used to evaluate geometric gaze estimation algorithms, allowing individual parts of an eye tracking system to be evaluated separately.
For example,~\citet{swirski2014rendering} used a rigged head model and reduced eyeball model to render ground truth images for evaluating pupil detection and tracking algorithms.

\begin{figure*}
    \includegraphics[width=\textwidth]{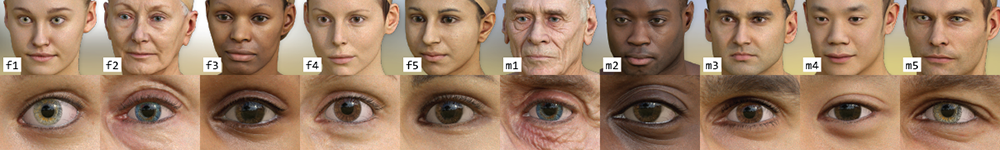}
    \caption{Our collection of head models and corresponding close-ups of the eye regions. The set exhibits a good range of variation in eye shape, surrounding bone structure, skin smoothness, and skin color.}
    \label{fig:model_suite}
\end{figure*}

\section{Dynamic Eye-Region Model}

We developed a realistic dynamic eye-region model which can be randomly posed to generate fully labeled training images.
Our goals were realism and controllability, so we combined 3D head scan geometry with our own posable eyeball model -- \autoref{fig:process} provides an overview of the model preparation process.
For the resulting training data to be useful, it should be representative of real-world variety.
We therefore aimed to model the continuous changes in appearance that the face and eyes undergo during eye movement, so they are accurately represented in close-up synthetic eye images.
This is more challenging than simply rendering a collection of static models, as dynamic geometry must be correctly topologized and rigged to be able to deform continuously.
Next, we present our anatomically inspired eyeball model and the procedure for converting a collection of static 3D head scans into dynamic eye-region models.

\subsection{Simplified Eyeball Model}
\label{subsec:eyeball_model}

\begin{figure}
    \captionsetup[subfigure]{labelformat=empty} % stop subcaption writing "(a)""
    \captionsetup{subrefformat=parens} % add parentheses to \subref
    \begin{subfigure}[t]{0.33\columnwidth}
        \inlinelabel{a}{\includegraphics[width=\textwidth]{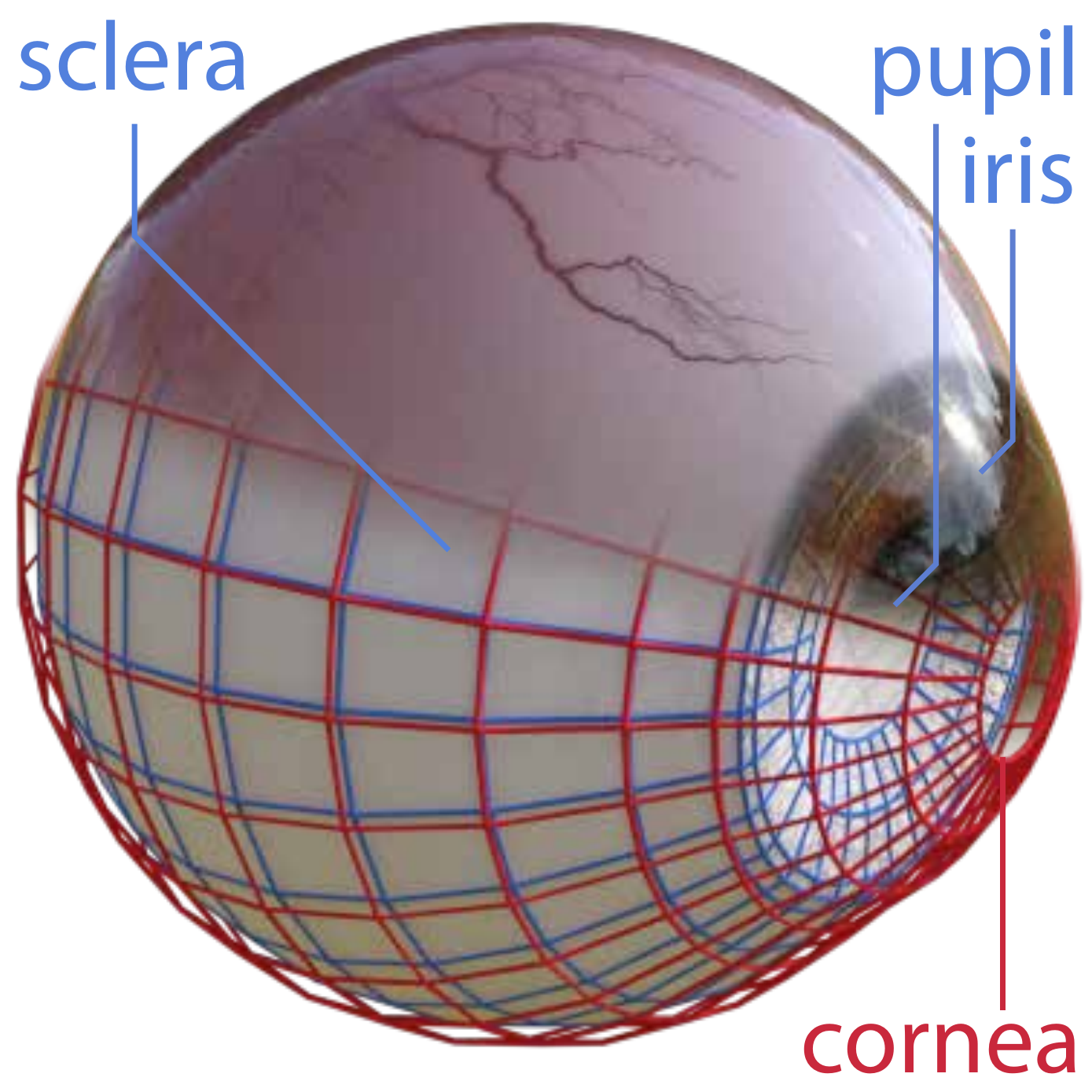}}
        \caption{}\label{fig:eye_model_parts}
    \end{subfigure}
    \hfill
    \begin{subfigure}[t]{0.65\columnwidth}
        \inlinelabel{b}{\includegraphics[width=\textwidth]{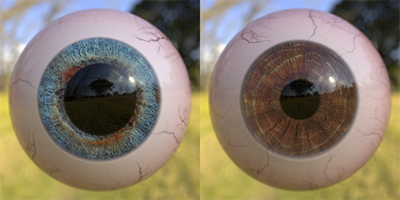}}
        \caption{}\label{fig:eye_model_images}
    \end{subfigure}
    \par\vspace{-28pt}
    \caption{Our eye model includes the sclera, pupil, iris, and cornea \subref{fig:eye_model_parts} and can exhibit realistic variation in both shape (pupillary dilation) and texture (iris color, scleral veins) \subref{fig:eye_model_images}.}
    \label{fig:eye_model}
\end{figure}

Our eye model consists of two parts (see~\autoref{fig:eye_model_parts}).
The outer part (red wireframe) approximates the eye's overall shape with two spheres ($r_1\!=\!12\textrm{mm}, r_2\!=\!8\textrm{mm}$ \cite{ruhland2014look}), the latter representing the corneal bulge.
To avoid a discontinuous seam between spheres, their meshes were joined, and the vertices along the seam were smoothed to minimize differences in face-angle.
This outer part is transparent, refractive ($n\!=\!1.376$), and partially reflective.
The sclera's bumpy surface is modeled with smoothed solid noise functions, and applied using a \emph{displacement map} -- a 2D scalar function that shifts a surface in the direction of its normal \cite{lee2000displaced}.
The inner part (blue wireframe) is a flattened sphere  -- the planar end represents the iris and pupil, and the rest represents the sclera, the white of the eye.
There is a $0.5\textrm{mm}$ gap between the two parts which accounts for the thickness of the cornea.

Eyes vary in both shape (pupillary dilation) and texture (iris color and scleral veins).
To model shape variation we use \emph{blend shapes}
to interpolate between several different poses created for the same topological mesh~\cite{orvalho2012facial}. 
We created blend shapes for dilated and constricted pupils, as well as large and small irises to account for a small amount ($10\%$) of variation in iris size.
We vary the texture of the eye by compositing images in three separate layers:
\begin{inparaenum}[\itshape i\upshape)]
\item a \emph{sclera} tint layer (white, pink, or yellow);
\item an \emph{iris} layer with four different photo-textures (amber, blue, brown, grey); and
\item a \emph{veins} layer (blood-shot or clear).
\end{inparaenum}

\subsection{3D Head Scan Acquisition}
\label{sec:eye_region_geom_prep}

For an eye-region rendering to be realistic, it must also feature realistic nearby facial detail.
While previous approaches used lifelike artist-created models \cite{swirski2014rendering}, we rely on high-quality head scans captured by a professional photogrammetry studio (10K diffuse color textures, 0.1mm resolution geometry)\footnote{Ten24 3D Scan Store -- \url{http://www.3dscanstore.com/}}.
Facial appearance around the eye varies dramatically between people as a result of different eye-shapes (e.g. round vs hooded), orbital bone structure (e.g. deep-set vs protruding), and skin detail (wrinkled vs smooth).
Therefore our head models (see \autoref{fig:model_suite}) cover gender, ethnicity and age.
As can be seen in \autoref{fig:process_original_scan}, the cornea of the original head scan has been incorrectly reconstructed by the optical scanning process.
This is because transparent surfaces are not directly visible, so cannot be reconstructed in the same way as diffuse surfaces, such as skin.
For images to represent a wide range of gaze directions, the eyeball needed to be posed separately from the face geometry.
We therefore removed the scanned eyeball from the mesh, and placed our own eyeball approximation in its place.

\subsection{Eye-Region Geometry Preparation}

While the original head scan geometry is suitable for being rendered as a static model, its high resolution topology cannot be easily controlled for changes in eye-region shape.
Vertical saccades are always accompanied by eyelid motion, so we need to control eyelid positions according to the gaze vector.
To do this, we need a more efficient (low-resolution) geometric representation of the eye-region, where edge loops flow around the natural contours of facial muscles.
This leads to more realistic animation as mesh deformation matches that of actual skin tissue and muscles \cite{orvalho2012facial}.

We therefore \emph{retopologized} the face geometry
using a commercial semi-automatic system\footnote{ZBrush ZRemesher 2.0, Pixologic, 2015}.
As can be seen in \autoref{fig:process_retopo}, this way edge loops followed the exterior eye muscles, allowing for realistic eye-region deformations.
This retopologized low-poly mesh lost the skin surface detail of the original scan, like wrinkles and creases (see \autoref{fig:process_displaced_subdiv}).
These were restored with a displacement map computed from the scanned geometry \cite{lee2000displaced}.
Although they are two separate organs, there is normally no visible gap between eyeball and skin.
However, as a consequence of removing the eyeball from the original scan, the retopologized mesh did not necessarily meet the eyeball geometry (see \autoref{fig:process_retopo}).
To compensate for this, the face mesh's eyelid vertices were automatically displaced along their normals to their respective closest positions on the eyeball geometry (see \autoref{fig:process_displaced_subdiv}).
This prevented unwanted gaps between the models, even after changes in pose.
The face geometry was then assigned physically-based materials, including subsurface scattering to approximate the penetrative light transfer properties of skin, and a glossy component to simulate its oily surface.

\subsection{Modeling Eyelid Motion and Eyelashes}

\begin{figure}
    \includegraphics[width=\columnwidth]{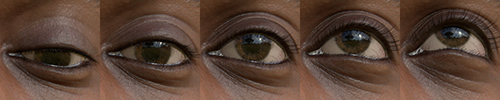} \par \smallskip
    \caption{Eyelids are posed by interpolating between blend shapes based on gaze direction (\texttt{m2} as example).}
    \label{fig:eyelids}
\end{figure}

We model eyelid motion using blend shapes for upwards-looking and downwards-looking eyelids, and interpolating between them based on the global pitch of the eyeball model.
This makes our face-model dynamic, allowing it to continuously deform to match eyeball poses.
Rather than rendering a single or perhaps several discrete head scans representing a particular gaze vector \cite{sugano2014learning}, we can instead create training data with a dense distribution of facial deformation.
Defining blend shapes through vertex manipulation can be a difficult and time-consuming task but fortunately, only two are required and they have small regions of support.
As the tissue around the eye is compressed or stretched, skin details like wrinkles and folds are either attenuated or exaggerated (see \autoref{fig:eyelids}).
We modeled this by using smoothed color and displacement textures for downwards-looking eyelids, removing any wrinkles.
These blend shape and texture modifications were carried out using photos of the same heads looking up and down as references.

Eyelashes are short curved hairs that grow from the edges of the eyelids.
These can occlude parts of the eye and affect eye tracking algorithms, so are simulated as part of our comprehensive model.
We followed the approach of {\'S}wirski and Dodgson~\cite{swirski2014rendering}, and modeled eyelashes using directed hair particle effects.
Particles were generated from a control surface manually placed below the eyelids.
To make them curl, eyelash particles experienced a slight amount of gravity during growth (negative gravity for the upper eyelash).

\section{Training Data Synthesis}

\begin{figure}
	\captionsetup[subfigure]{labelformat=empty} % stop subcaption writing "(a)""
    \captionsetup{subrefformat=parens} % add parentheses to \subref
    \centering
    \begin{subfigure}[t]{0.48\columnwidth}
        \inlinelabel{a}{\includegraphics[width=\textwidth]{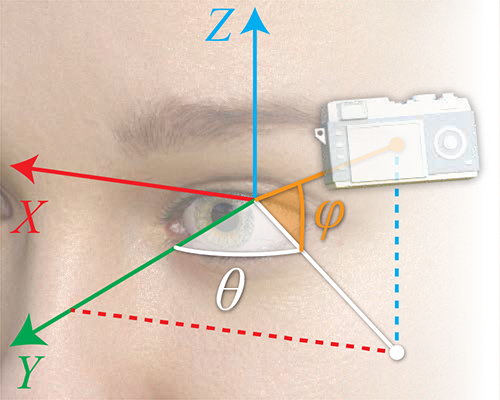}}
        \caption{}\label{fig:cam_pos_spher_coords}
    \end{subfigure}
    \hfill
    \begin{subfigure}[t]{0.48\columnwidth}
        \inlinelabel{b}{\includegraphics[width=\textwidth]{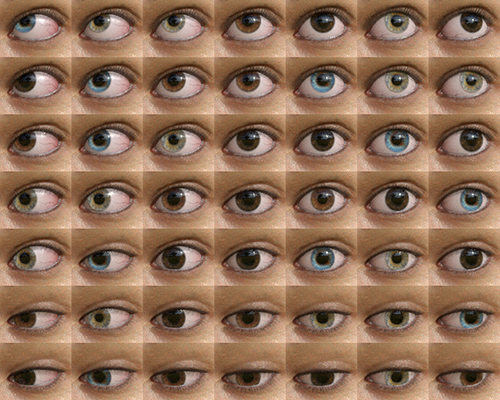}}
        \caption{}\label{fig:cam_pos_example_renders}
    \end{subfigure}
    \par\vspace{-28pt}
    \caption{The camera is positioned to simulate changes in head pose \subref{fig:cam_pos_spher_coords}. At each position, we render many eye images for different gaze directions by posing the eyeball model \subref{fig:cam_pos_example_renders}.}
    \label{fig:cam_pos}
\end{figure}

In-the-wild images exhibit large amounts of appearance variability across different viewpoints and illuminations.
Our goal was to sufficiently sample our model across these degrees of variation to create representative image datasets.
In this section we first describe how we posed our viewpoint and model, and explain our approach for using image-based lighting \cite{debevec2002image} to model a wide range of realistic environments.
We then describe our landmark annotation process and finally discuss the details of our rendering setup.

\subsection{Posing the Model}

For a chosen eye-region model, each rendered image is determined by parameters $(\mathbf{c}, \mathbf{g}, L, E)$: 3D camera position $\mathbf{c}$; 3D gaze vector $\mathbf{g}$; lighting environment $L$; and eye model configuration $E$.
Camera positions $\mathbf{c}$ were chosen by iterating over spherical coordinates $(\theta, \phi)$, centered around the eyeball center (see~\autoref{fig:cam_pos}).
We used orthographic rendering, as this simulates an eye region-of-interest being cropped from a wide-angle camera image.
At each camera position $\mathbf{c}$, we rendered multiple images with different 3D gaze vectors to simulate the eye looking in different directions.
Examples with fixed $L$ are shown in \autoref{fig:cam_pos_example_renders}.
Gaze vectors $\mathbf{g}$ were chosen by first pointing the eye directly at the camera (simulating eye-contact), and then modifying the eyeball's pitch ($\alpha$) and yaw ($\beta$) angles over a chosen range.
Within $E$ we randomly configure iris color and pose eyelids according to $\mathbf{g}$.
For our generic dataset, we rendered images with up to $45^{\circ}$ horizontal and vertical deviation from eye-contact, in increments of $10^{\circ}$.
As we posed the model in this way, there was the possibility of rendering ``unhelpful'' images that either simulate impossible scenarios or are not useful for training.
To avoid violating anatomical constraints, we only rendered images for valid eyeball rotations $|\alpha|\!\leq\!25^{\circ}$ and $|\beta|\!\leq\!35^{\circ}$ \cite{MIL-STD-1472G}.
Before rendering, we also verified that the projected 2D pupil center in the image was within the 2D boundary of the eyelid landmarks -- this prevented us from rendering images where too little of the iris was visible.

\subsection{Creating Realistic Illumination}

\begin{figure}
    \begin{subfigure}[t]{\columnwidth}
        \includegraphics[width=0.24\textwidth]{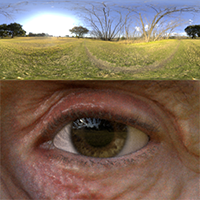} \hfill
    	\includegraphics[width=0.24\textwidth]{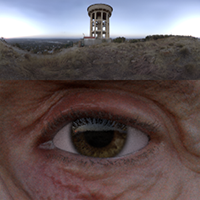} \hfill
        \includegraphics[width=0.24\textwidth]{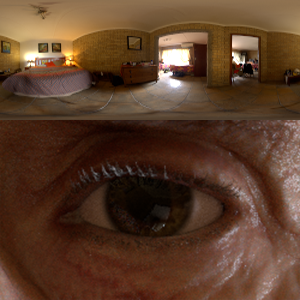} \hfill
    	\includegraphics[width=0.24\textwidth]{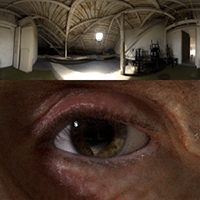}
	    \caption{The four HDR environment maps we use for realistic lighting: bright/cloudy outdoors, and bright/dark indoors}
    \end{subfigure}
    \par \medskip
    \begin{subfigure}[t]{0.48\columnwidth}
        \includegraphics[width=\textwidth]{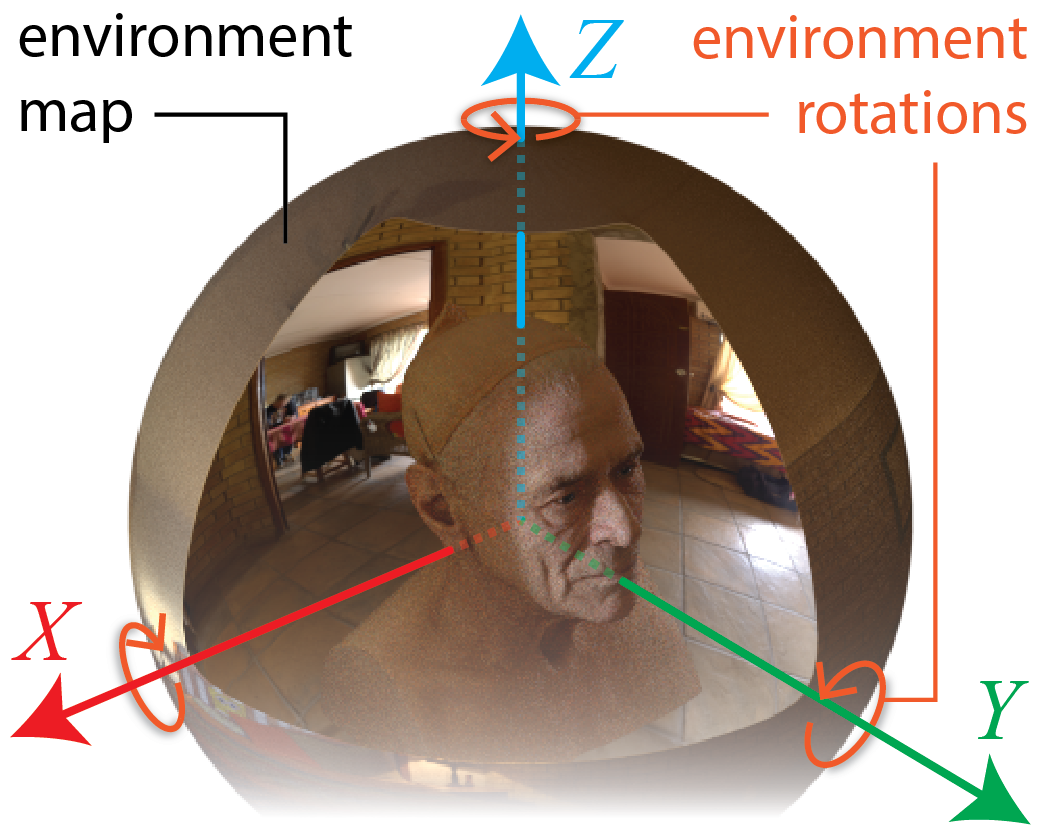}
    	\caption{The environment is rotated to simulate different head poses}
    \end{subfigure}%
    \hfill
    \begin{subfigure}[t]{0.48\columnwidth}
        \includegraphics[width=\textwidth]{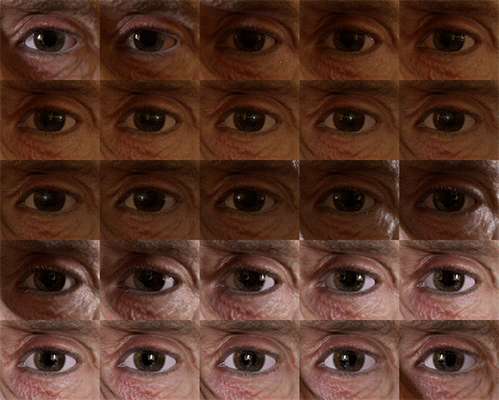}
        \caption{Renders using a single environment, rotated about $Z$}
        \label{fig:env_map_imgs_examples}
    \end{subfigure}
    \caption{Appearance variation from lighting is modelled with poseable high dynamic range environment maps \cite{debevec2002image}.}
    \label{fig:environment_maps}
\end{figure}

One of the main challenges in computer vision is illumination invariance -- a good system should work under a range of real-life lighting conditions.
We realistically illuminate our eye-model using \emph{image-based lighting}, a technique where high dynamic range (HDR) panoramic images are used to provide light in a scene \cite{debevec2002image}.
This works by photographically capturing omni-directional light information, storing it in a texture, and then projecting it onto a sphere around the object.
When a ray hits that texture during rendering, it takes that texture's pixel value as light intensity.
At render time we randomly chose one of four freely available HDR environment images\footnote{\url{http://adaptivesamples.com/category/hdr-panos/}} to simulate a range of different lighting conditions (see \autoref{fig:environment_maps}).
The environment is then randomly rotated to simulate a continuous range of head-pose, and randomly scaled in intensity to simulate changes in ambient light.
As shown in \autoref{fig:env_map_imgs_examples}, a combination of hard shadows and soft light can generate a range of appearances from only a single HDR environment.

\subsection{Eye-Region Landmark Annotation}

For eye shape registration, we needed additional ground-truth annotations of eye-region landmarks in the training images.
As shown in \autoref{fig:process_ldmks}, each 3D eye-region was annotated once in 3D with $28$ landmarks, corresponding to the eyelids ($12$), iris boundary ($8$), and pupil boundary ($8$).
The iris and pupil landmarks were defined as a subset of the eyeball geometry vertices, so deform automatically with changes in pupil and iris size.
The eyelid landmarks were manually labelled with a separate mesh that follows the seam where eyeball geometry meets skin geometry.
This mesh is assigned shape keys and deforms automatically during eyelid motion.
Whenever an image is rendered, the 2D image-space coordinates of these 3D landmarks are calculated using the camera projection matrix and saved.

\subsection{Rendering Images}

We use Blender's\footnote{The Blender Project -- \url{http://www.blender.org/}} inbuilt Cycles path-tracing engine for rendering.
This Monte Carlo method traces the paths of many light rays per pixel, scattering light stochastically off physically-based materials in the scene until they reach illuminants.
A GPU implementation is available for processing large numbers of rays simultaneously ($150/\textrm{px}$) to achieve noise-free and photorealistic images.
We rendered a generic \dataset dataset of 11,382 images covering $40^{\circ}$ of viewpoint (i.e. head pose) variation and $90^{\circ}$ of gaze variation.
We sampled eye colour and environmental lighting randomly for each image.
Each $120\!\times\!80\textrm{px}$ rendering took $5.26\textrm{s}$ on average using a commodity GPU (Nvidia GTX660).
As a result we can specify and render a cleanly-labelled dataset in under a day on a single machine -- a fraction of the time taken by traditional data collection procedures \cite{zhang15_cvpr}.

\section{Experiments}

We evaluated the usefulness of our synthetic data generation method on two sample problems, eye-shape registration and appearance-based gaze estimation.

Eye-shape registration attempts to detect anatomical landmarks of the eye -- eyelids, iris and the pupil. 
Such approaches either attempt to model the shape of the eye directly by relying on low-level image features, e.g. edges \cite{wood2014eyetab, swirski2012robust} or by using statistically learnt deformable models \cite{alabort2014statistically}. 
Compared to \citet{alabort2014statistically}, our dataset has been automatically labelled. This guarantees consistent labels across viewpoints and people, avoiding human error.

Appearance-based gaze estimation systems learn a mapping directly from eye image pixels to gaze direction.
While most previous approaches focused on {\em person-dependent} training scenarios which require training data from the target user, recently more attention has been paid to {\em person-independent} training \cite{zhang15_cvpr,sugano2014learning,funes2013person,schneider2014manifold}. 
The training dataset is required to cover the potential changes in appearance with different eye shapes, arbitrary head poses, gaze directions, and illumination conditions.
Compared to \citet{sugano2014learning}, our method can provide a wider range of illumination conditions which can be beneficial to handle the unknown illumination condition in the target domain.

\subsection{Eye-Shape Registration}

As our method can reliably generate consistent landmark location training data, we used it for training a Constrained Local Neural Field (CLNF) \cite{baltrusaitis2013constrained} deformable model. We conducted experiments to evaluate the generalizability of our approach on two different use cases: eyelid registration in-the-wild, and iris tracking from webcams.

\paragraph{Eyelid Registration In the Wild}

\begin{figure}
    \centering
    \includegraphics[width=\columnwidth]{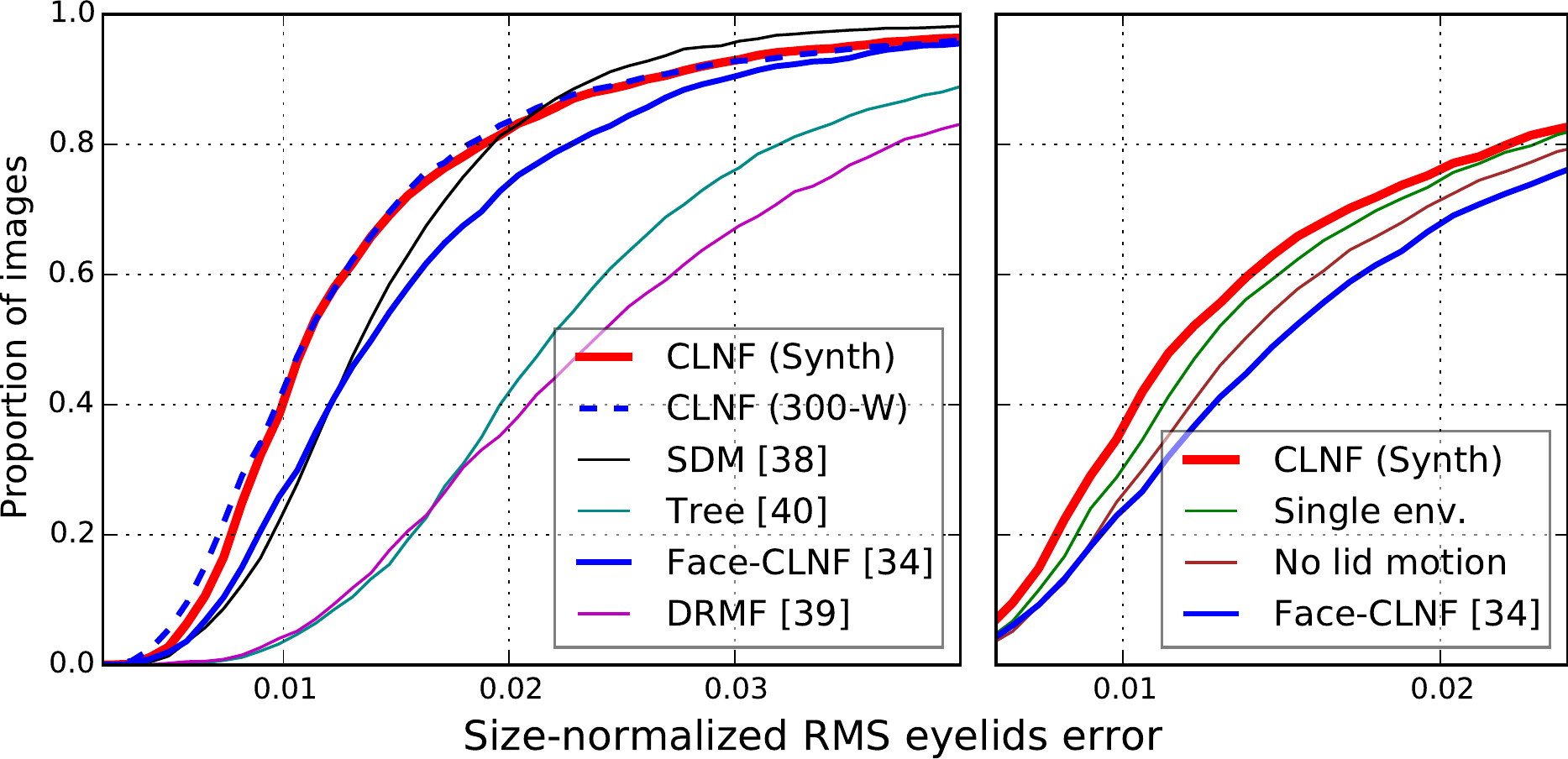}
    \caption{We outperform the state-of-the-art for eyelid-registration in the wild. The right plot shows how performance degrades for training data without important degrees of variation: realistic lighting and eyelid movement.}
    \label{fig:clnf_results_wild}
\end{figure}

\begin{figure}
    \centering
    \includegraphics[width=\columnwidth]{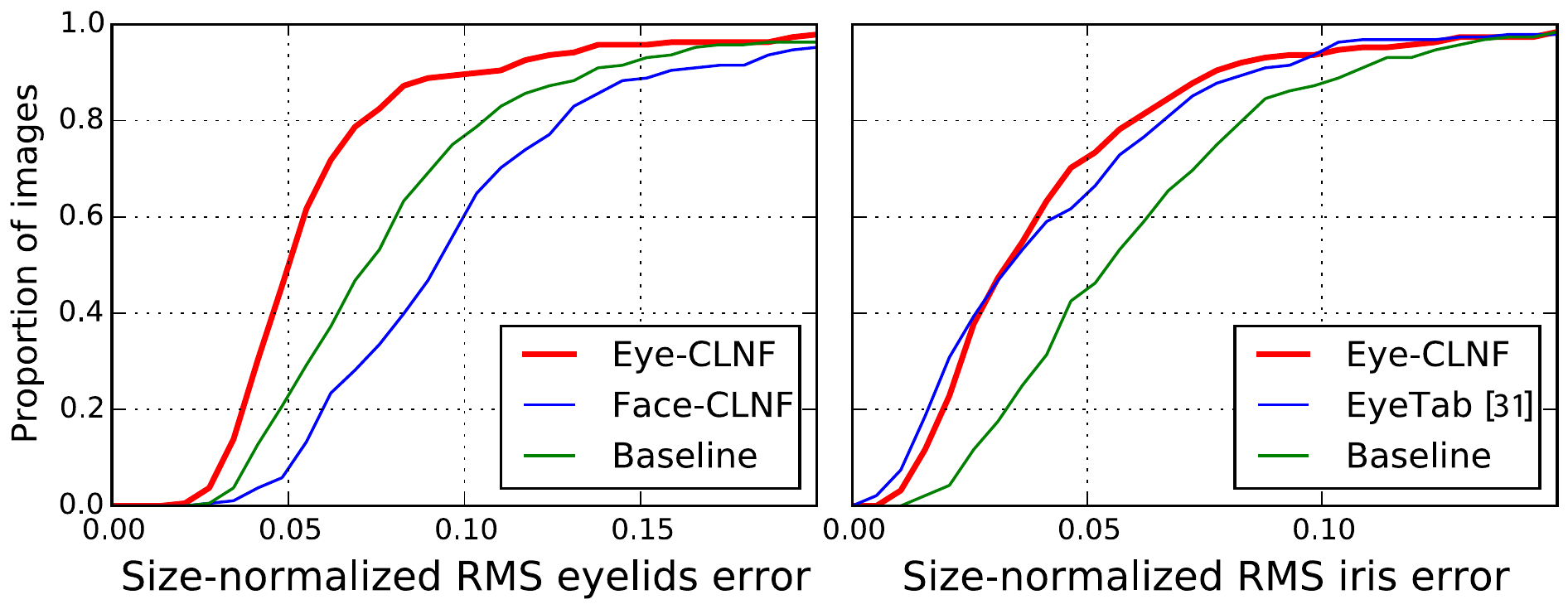}
    \caption{We perform comparably with state-of-the-art for iris-registration on in-the-wild webcam images.}
    \label{fig:clnf_results_MPII}
\end{figure}

We performed an experiment to see how our system generalizes on unseen and unconstrained images. We used the validation datasets from the 300 Faces In-the-Wild (300-W) challenge \cite{sagonas2013300} which contain labels for eyelid boundaries. We tested all of the approaches on the 830 (out of 1026) test images. We discarded images that did not contain visible eyes (occluded by hair or sunglasses) or where face detection failed for other comparison systems used in our experiment.

We trained CLNF patch experts using the generic \dataset dataset and used the 3D landmark locations to construct a Point Distribution Model (PDM) using Principal Component Analysis. 
As our rendered images did not contain closed eyes we generated extra closed eye landmark labels by moving the upper eyelid down to lower one or meeting both eyelids halfway.
We initialized our approach by using the face-CLNF \cite{baltrusaitis2013constrained} facial landmark detector.
To compare using synthetic or real training images, we trained an eyelid CLNF model on 300-W images, but used the same PDM used for synthetic data (CLNF 300-W).
We also compared our approach with the following state-of-the-art facial landmark detectors trained on in-the-wild data: CLNF \cite{baltrusaitis2013constrained}, Supervised Descent Method (SDM) \cite{Xiong2013sdm}, Discriminative Response Map Fitting (DRMF) \cite{Asthana2013drmf}, and tree based face and landmark detector \cite{Zhu2012tree}. 

\begin{figure*}[ht]
    \captionsetup[subfigure]{labelformat=empty} % stop subcaption writing "(a)""
    \captionsetup{subrefformat=parens} % add parentheses to \subref
    \centering
    \begin{subfigure}[t]{\columnwidth}
        \inlinelabel{a}{\includegraphics[width=\textwidth]{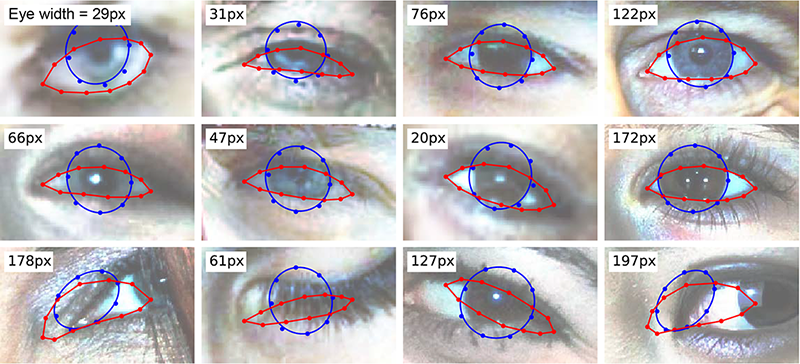}}
        \caption{}\label{fig:fits_300W}
    \end{subfigure}
    \hfill
    \begin{subfigure}[t]{\columnwidth}
        \inlinelabel{b}{\includegraphics[width=\textwidth]{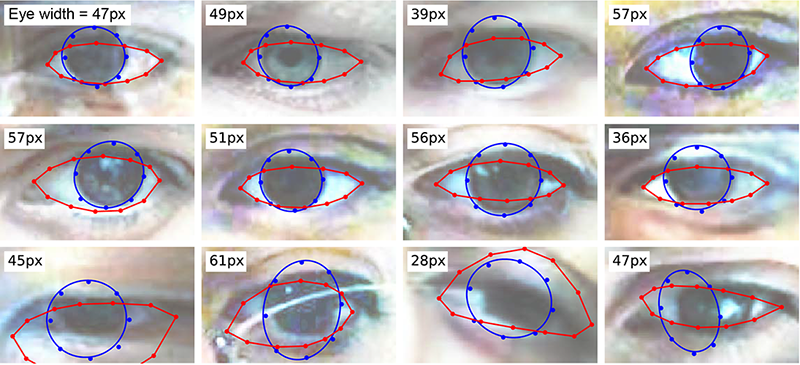}}
        \caption{}\label{fig:fits_MPII}
    \end{subfigure}
    \par\vspace{-28pt}
    \caption{Example fits of our \dataset eye-CLNF on in-the-wild images \subref{fig:fits_300W} and webcam images \subref{fig:fits_MPII}. The top two rows illustrate successful eye-shape registrations, while the bottom row illustrates failure cases, including unmodelled occlusions (hair), unmodelled poses (fully closed eye), glasses, and incorrect model initialization. Note our algorithm generalizes well to eye images of different sizes.}
    \label{fig:example_CLNF_fits}
\end{figure*}

The results of our experiments can be seen in \autoref{fig:clnf_results_wild}, and example model fits are shown in \autoref{fig:fits_300W}.
Errors were recorded as the RMS point-to-boundary distance from tracked eyelid landmarks to ground truth eyelid boundary, and were normalized by inter-ocular distance. 
First, our system CLNF Synth ($\mathrm{Mdn}=0.0110$) trained on only 10 participants in four lighting conditions results in very similar performance to a system trained on unconstrained in-the-wild images, CLNF 300-W ($\mathrm{Mdn}=0.0110$).
Second, the results show the eye-specific CLNF outperformed all other systems in eye-lid localization: SDM ($\mathrm{Mdn}=0.0134$), face-CLNF ($\mathrm{Mdn}=0.0139$), DRMF ($\mathrm{Mdn}=0.0238$), and Tree based ($\mathrm{Mdn}=0.0217$). 
The first result suggests the importance of high-quality consistent labels. In addition, we perform well despite the fact our models do not exhibit emotion-related shape deformation, such as brow-furrowing, squinting, and eye-widening.

Our approach also allow us to examine what steps of the synthesis are important for generating good training data. We trained two further eye-specific CLNFs on different versions of \dataset, one without eyelid motion and one with only one fixed lighting condition. As can be seen in \autoref{fig:clnf_results_wild}, not using shape variation ($\mathrm{Mdn}=0.0129$) and using basic lighting ($\mathrm{Mdn}=0.0120$) lead to worse performance due to missing degrees of variability in training sets.

\paragraph{Eye-Shape Registration for Webcams}

%!TEX root = ../00_main.tex

%

While the 300-W images represent challenging conditions for eyelid registration they do not feature iris labels and are not representative of conditions encountered during everyday human-computer interaction.
We therefore annotated sub-pixel eyelid and iris boundaries for a subset of MPIIGaze \cite{zhang15_cvpr} (188 images), a recent large-scale dataset of face images and corresponding on-screen gaze locations collected during everyday laptop use over several months \cite{zhang15_cvpr}.
Pupil accuracy was not evaluated as it was impossible to discern in in most images.

We compared our eye-specific CLNF (CLNF Synth) with EyeTab \cite{wood2014eyetab}, a state-of-the-art shape-based approach for webcam gaze estimation that robustly fits ellipses to the iris boundary using image-aware RANSAC \cite{swirski2012robust}. Note we did not compare with other systems from the previous experiment as they do not detect irises.
We used a modified version of the author's implementation with improved eyelid localization using CLNF \cite{baltrusaitis2013constrained}.
As a baseline, we used the mean position of all 28 eye-landmarks following model initialization.
Eyelid errors were calculated as RMS distances from predicted landmarks to the eyelid boundary.
Iris errors were calculated by least-squares fitting an ellipse to the tracked iris landmarks, and measuring distances only to visible parts of the iris.
Errors were normalized by the eye-width, and are reported using average eye-width ($44.4\textrm{px}$) as reference.

As shown in \autoref{fig:clnf_results_MPII}, our approach ($\textrm{Mdn}\!=\!1.48\textrm{px}$) demonstrates comparable iris-fitting accuracy with EyeTab ($\textrm{Mdn}\!=\!1.44\textrm{px}$).
However, CLNF Synth is more robust, with EyeTab failing to terminate in $2\%$ of test cases.
As also shown by the 300-W experiment, the eye-specific CLNF Synth localizes eyelids better than the face-CLNF.
See \autoref{fig:fits_MPII} for example model fits.

\subsection{Appearance-Based Gaze Estimation}

\begin{figure}
    \centering
    \includegraphics[width=0.9\columnwidth]{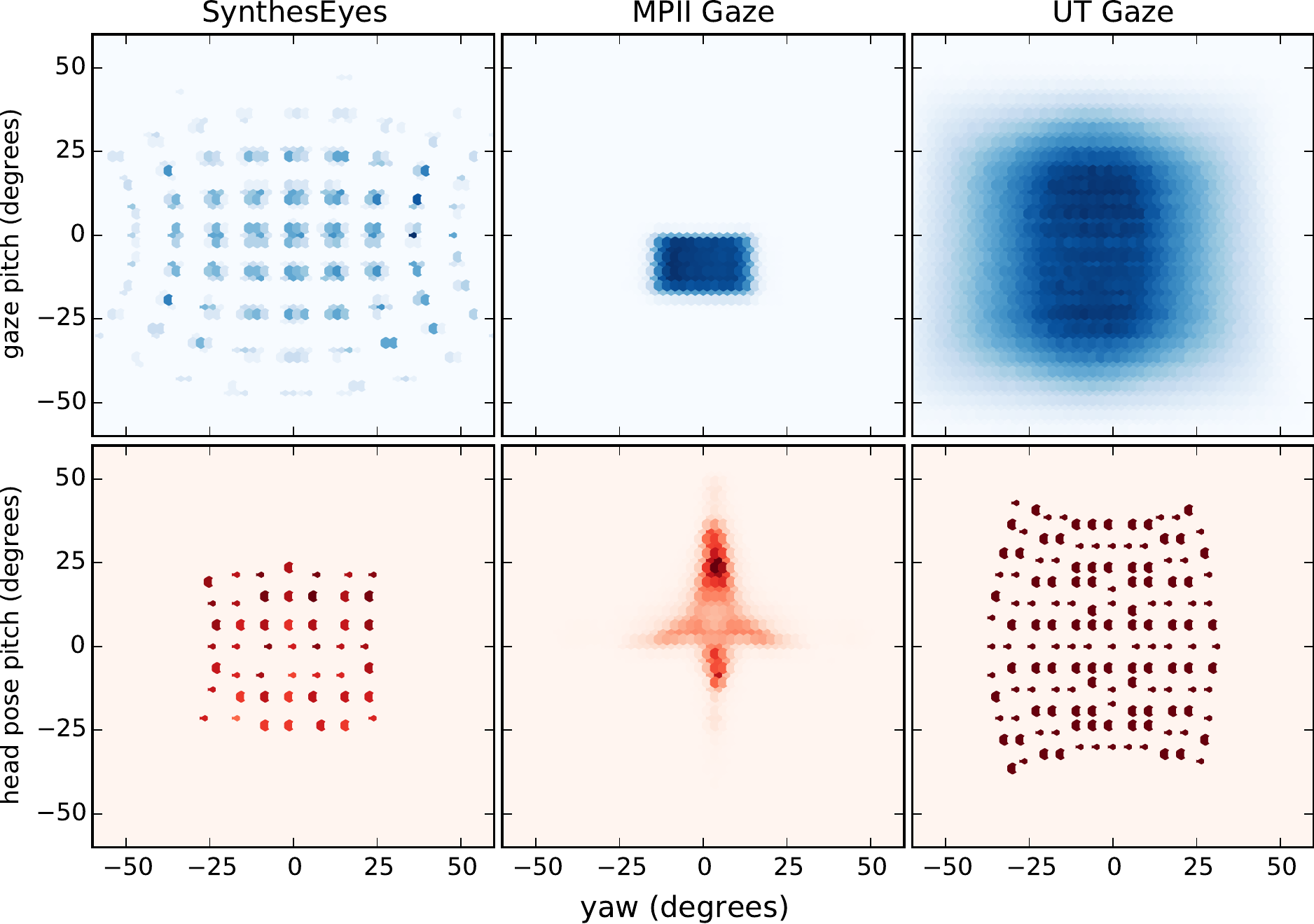}
    \caption{The gaze direction (first row) and head pose (second row) distributions of different datasets: \dataset, MPIIGaze~\cite{zhang15_cvpr}, and UT Multiview \cite{sugano2014learning}.}
    \label{fig:head_gaze_distribution}
\end{figure}

\begin{figure}
    \centering
    \includegraphics[width=\columnwidth]{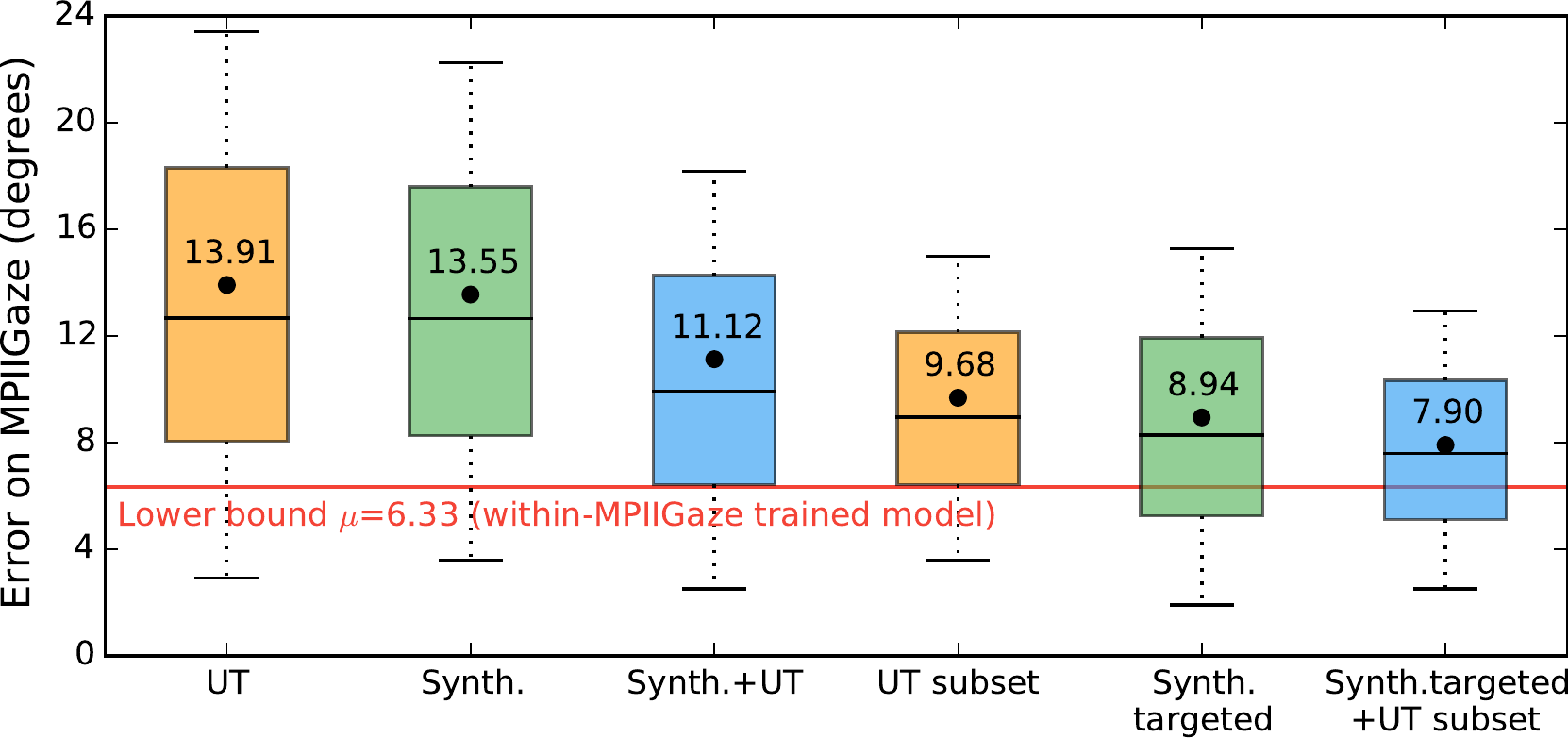}
    \caption{Test performance on MPIIGaze; x-axis represents training set used. Dots are mean errors, and red line represents a practical lower-bound (within-dataset cross-validation score). Note how combining synthetic datasets for training lead to improved performance (blue plots).}
    \label{fig:gazeResult}
\end{figure}

To evaluate the suitability of our synthesis method for appearance-based gaze estimation we performed a cross-dataset experiment as described by \citet{zhang15_cvpr}.
We synthesized training images using the same camera settings as in the UT dataset~\cite{sugano2014learning}.
The head pose and gaze distributions for the three datasets are shown in~\autoref{fig:head_gaze_distribution}.
We then trained the same convolutional neural network (CNN) model as in~\cite{zhang15_cvpr} on both synthetic datasets and evaluated their performance on MPIIGaze.
As shown
in~\autoref{fig:gazeResult}, the CNN model trained on our generic~\dataset dataset achieved similar performance ($\mu\!=\!13.91^{\circ}$) as the model trained on the UT dataset \mbox{($\mu\!=\!13.55^{\circ}$)}.
This confirms that our approach can synthesize data that leads to comparable results with previous synthesis procedures \cite{sugano2014learning}.
Note from \autoref{fig:gazeResult} that there is still a performance gap between this cross-dataset and the within-dataset training (red line).

While it is in general important to cover a wide range of head poses to handle arbitrary camera settings
, if the target setting is known in advance, e.g. laptop gaze interaction as in case of MPIIGaze, it is possible to target data synthesis to the expected head pose and gaze ranges.
To study the ability of our method to perform such a targeting, we rendered an additional dataset (\dataset targeted) for a typical laptop setting ($10^{\circ}$ pose and $20^{\circ}$ gaze variation).
For comparison, we also re-sampled the entire UT dataset to create a subset (UT subset) that has the same gaze and head pose distribution as MPIIGaze.
To make a comparison assuming the same number of participants, we further divided the UT subset into five groups with 10 participants each,
and averaged the performance of the five groups for the final result. 
As shown in the third and forth bars of~\autoref{fig:gazeResult}, having similar head pose and gaze ranges as the target domain improves performance compared to the generic datasets.
Trained on our \dataset dataset the CNN achieves a statistically significant performance improvement over the UT dataset of 0.74$^{\circ}$ (Wilcoxon signed-rank test: $p\!<\!0.0001$).

These results suggest that neither \dataset~nor the UT dataset alone capture all variations present in the test set, but different ones individually.
For example, while we cover more variations in lighting and facial appearance, the UT dataset contains real eye movements captured from more participants.
Recent works by \citet{fu2011neural} and \citet{peng2014exploring} demonstrated the importance of fine-tuning models initially trained on synthetic data on real data to increase performance.
Finally, we therefore evaluated the performance by training and fine-tuning using both datasets (see~\autoref{fig:gazeResult}).
We first trained the same CNN model on the~\dataset~dataset and fine-tuned the model using the UT dataset.
This fine-tuned model achieved better performances in both settings (untargeted $\mu\!=\!11.12^{\circ}$, targeted $\mu\!=\!7.90^{\circ}$).
The performance of the untargeted case significantly outperformed the state-of-the-art result~\cite{zhang15_cvpr} (Wilcoxon signed-rank test: $p\!<\!0.0001$), and indicates a promising way for a future investigation to fill the performance gap.

\begin{figure}
    \centering
    \includegraphics[width=\columnwidth]{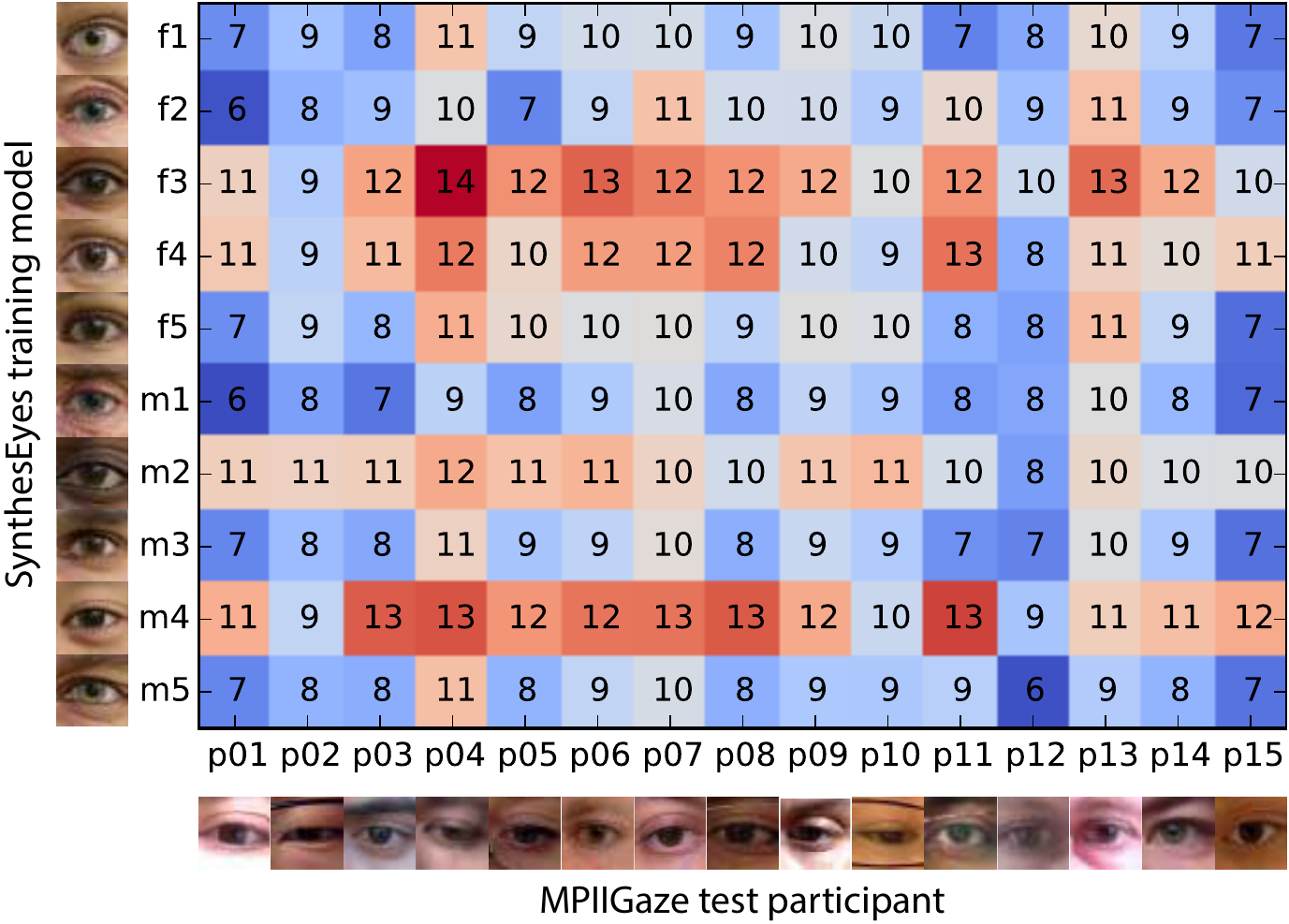}
    \caption{Per--eye-model gaze estimation mean errors on MPIIGaze. Red represents worst scores. Note how some eye-models have proved more useful than others for training.
    }
    \label{fig:person_specific_training}
\end{figure}

\paragraph{Person-Specific Appearance}

Appearance-based gaze estimation performs best when trained and tested on the same person, as the training data includes the same eye appearances that occur during testing.
However, eye images from \dataset and MPIIGaze can appear different due to differences in eye-shape and skin color.
To examine the effects of this we conducted a second experiment where we trained 10 separate systems (one trained on each \dataset eye model) and tested on each participant in MPIIGaze.
The results can be seen in \autoref{fig:person_specific_training}.

This plot illustrates which \dataset models were useful for training and which ones were not.
As we can see, training with certain eye models lead to poor generalization, for example \texttt{f3}, \texttt{m2}, and \texttt{m4}, perhaps due to differences in skin-tone and eye-shape.
Also, total errors for some target participants are lower than for others, perhaps because of simpler eye-region shape that is matched to the training images.
Although intuitive, these experiments further confirm the importance of correctly covering appearance variations in the training data.
They also open up potential directions for future work, including person-specific adaptation of the renderings and gaze estimation systems.

\section{Conclusion}

We presented a novel method to synthesize perfectly labelled realistic close-up images of the human eye.
At the core of our method is a computer graphics pipeline that uses a collection of dynamic eye-region models obtained from head scans to generate images for a wide range of head poses, gaze directions, and illumination conditions.
We demonstrated that our method outperforms state-of-the-art methods for eye-shape registration and cross-dataset appearance-based gaze estimation in the wild.
These results are promising and underline the significant potential of such learning-by-synthesis approaches particularly in combination with recent large-scale supervised methods.

{\small
\bibliographystyle{IEEEtranN}
\bibliography{bib}
}

\end{document}